\documentclass[conference]{IEEEtran}
\IEEEoverridecommandlockouts
\usepackage{cite}
\usepackage{amsmath,amssymb,amsfonts}
\usepackage{algorithmic}
\usepackage{graphicx}
\usepackage{textcomp}
\usepackage{xcolor}
\def\BibTeX{{\rm B\kern-.05em{\sc i\kern-.025em b}\kern-.08em
    T\kern-.1667em\lower.7ex\hbox{E}\kern-.125emX}}

\usepackage{subfiles}
\usepackage{blindtext}

\usepackage{enumitem}

\ifCLASSOPTIONcompsoc
    \usepackage[caption=false, font=normalsize, labelfont=sf, textfont=sf]{subfig}
\else
\usepackage[caption=false, font=footnotesize]{subfig}
\fi

\usepackage[hidelinks]{hyperref}

\begin{document}

\title{Inferring Convolutional Neural Networks' accuracies from their architectural characterizations\\}

\author{\IEEEauthorblockN{1\textsuperscript{st} Duc Hoang}
\IEEEauthorblockA{\textit{Department of Physics} \\
\textit{Rhodes College}\\
Memphis, Tennessee 38112 USA \\
hoadm-21@rhodes.edu}
\and
\IEEEauthorblockN{2\textsuperscript{nd} Jesse Hamer}
\IEEEauthorblockA{\textit{Department of Mathematics} \\
\textit{The University of Iowa}\\
Iowa City, Iowa, 52242, USA \\
jhamer90811@gmail.com}
\and
\IEEEauthorblockN{3\textsuperscript{rd} Gabriel N. Perdue}
\IEEEauthorblockA{\textit{Quantum Science} \\
\textit{Fermi National Accelerator Laboratory (FNAL)}\\
Batavia, Illinois, 60510, USA \\
perdue@fnal.gov}
\and
\IEEEauthorblockN{4\textsuperscript{th} Steven R. Young}
\IEEEauthorblockA{\textit{Oak Ridge National Laboratory} \\
Oak Ridge, Tennessee, 37830, USA\\
youngsr@ornl.gov}
\and
\IEEEauthorblockN{5\textsuperscript{th} Jonathan Miller}
\IEEEauthorblockA{\textit{Universidad T\'ecnica Federico Santa Mar\'{\i}a} \\
Valpara\'iso, Chile\\
jonathan.miller@usm.cl}
\and
\IEEEauthorblockN{6\textsuperscript{th} Anushree Ghosh}
\IEEEauthorblockA{\textit{Universidad T\'ecnica Federico Santa Mar\'{\i}a} \\
Valpara\'iso, Chile\\
anushree.ghosh@usm.cl}
}

\maketitle

\begin{abstract}
Convolutional Neural Networks (CNNs) have shown strong promise for analyzing scientific data from many domains including particle imaging detectors.
However, the challenge of choosing the appropriate network architecture (depth, kernel shapes, activation functions, etc.) for specific applications and different data sets is still poorly understood.
In this paper, we study the relationships between a CNN's architecture and its performance by proposing a systematic language that is useful for comparison between different CNN's architectures before training time.
We characterize CNN's architecture by different attributes, and demonstrate that the attributes can be predictive of the networks’ performance in two specific computer vision-based physics problems --- event vertex finding and hadron multiplicity classification in the MINERvA experiment at Fermi National Accelerator Laboratory.
In doing so, we extract several architectural attributes from optimized networks’ architecture for the physics problems, which are outputs of a model selection algorithm called Multi-node Evolutionary Neural Networks for Deep Learning (MENNDL).
We use machine learning models to predict whether a network can perform better than a certain threshold accuracy before training.
The models perform 16-20\% better than random guessing.
Additionally, we found an coefficient of determination of 0.966 for an Ordinary Least Squares model in a regression on accuracy over a large population of networks.
\end{abstract}


\section{Introduction} \label{Introduction}
Deep Learning (DL) is a sub-field of machine learning that focuses on learning features from data through multiple layers of abstraction. Deep Convolutional Neural Networks (CNNs) have become the state-of-the-art DL technique in the fields of computer vision, natural language processing, and other scientific research domains such as High Energy Physics \cite{YLeCun-DL-Nature-2015,LSong-DL-2019}. 
That said, due to CNNs' inability to generalize for all datasets, a necessary step before applying CNNs to new data is selecting an appropriate set of architecture hyper-parameters. 
In the literature, the problem of choosing a CNN's architecture which is well suited to a given problem domain is still poorly understood. 
Generally, while there have been many studies of automated architecture search \cite{DBLP:journals/corr/abs-1905-01392}, very little has been done to develop a standardized language for describing neural network architectures in such a way as to be useful for comparison of multiple networks, or prediction of network performance metrics on the basis of architectural parameters. 
Many model selection algorithms have been proposed to mitigate the hyper-parameter optimization process, yet they mainly rely on human intuition or random search as of which parameter search space should be explored \cite{Bergstra:2012:RSH:2503308.2188395}.
In this study, we will thus propose a systematic language to characterize CNN architecture for simple, modular networks, and focus on demonstrating that different characterizations of the network architecture can be predictive of its performance in two computer vision problems in a particle physics context--vertex finding \cite{ijcnn7966131,Perdue_2018,LSong-DL-2019} and hadron multiplicity in MINERvA.
MINERvA \cite{Aliaga2014130} is a neutrino-nucleus scattering experiment at Fermi National Accelerator Laboratory with fine-grained, stereoscopic imaging capabilities and few-nanosecond timing resolution.
We conclude that our architectural attributes set can be used to give us partial insights into a network's performance prior to training.
We will also present specific architectural attributes that are highly relevant to CNNs' performance for those problems for further study and development of the models. 

In this work, \emph{network architecture} refers to the structural qualities of the network which are specified \emph{prior to training}: e.g. the layer types, the ordering of layer types, the layer non-linearities, and layer-specific parameters like the widths of fully-connected layers and the kernel shapes and strides of convolutional layers.
Network architecture can be considered as a subset of the network hyper-parameters, though in this work, we do not consider hyper-parameters such as learning rate, momentum, optimization method, or anything else related to the learning process.
We also do not include learned weights and biases as attributes characterizing network architecture, although it would be interesting to see how different architectures, trained on the same dataset and subject to the same loss functions, learn different feature maps for each of their layers.
All of the networks in each population presented here were trained with identical learning rate schedules and for an identical number of iterations.
Only the architecture was varied, so we use the term \emph{attribute} to denote architectural properties of neural networks.

\begin{figure}[tp]
\centering
\includegraphics[width=0.5\textwidth]{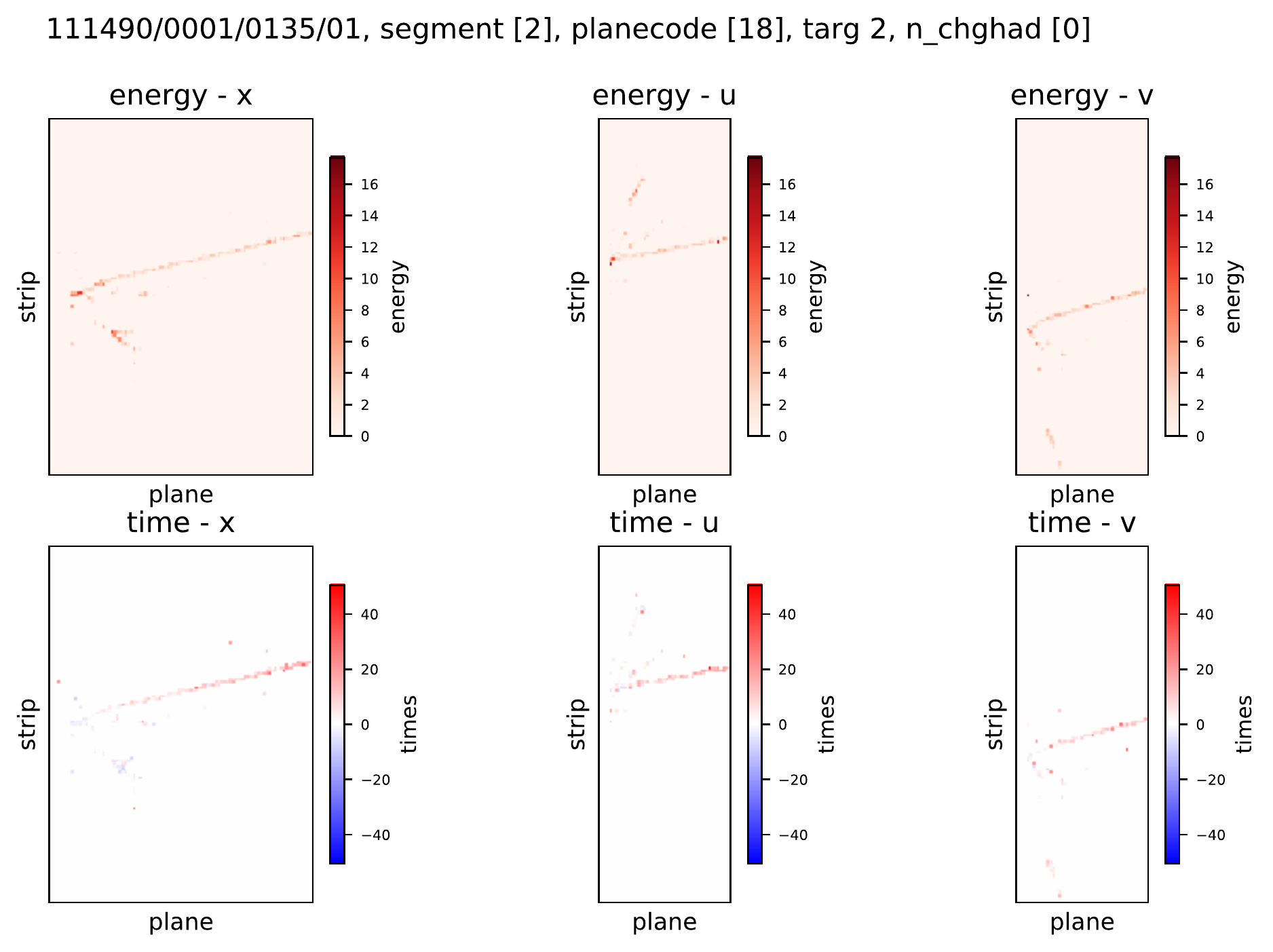}
\caption{\small A sample image taken by the MINERvA detector. 
Here we show a subset of the active channels and do not show structural elements.
Each image has three ``views'', labeled $x$, $u$, and $v$, with a relative time and deposited energy channel each, which together may be used to infer how the particle interaction occurred in 3D space.
The ``strip'' axis in the figure counts position along these planes, and the ``plane'' axis in the figure counts over the planes (stacked horizontally in roughly the direction of the neutrino beam).
Note that the views are actually interleaved into a repeating ``UXVX'' pattern, but they are grouped together by view for consumption by the CNN.
}\label{fig:minerva_input}
\end{figure}

The networks analyzed here are convolutional networks trained by an evolutionary algorithm called MENNDL (Multi-node Evolutionary Neural Networks for Deep Learning) \cite{Young:2015:ODL:2834892.2834896}.
The networks were trained for the task of vertex finding \cite{LSong-DL-2019} and hadron multiplicity counting in images collected from Fermilab's MINERvA detector\footnotemark.
\footnotetext{minerva.fnal.gov}
For the vertex finding task, in each image input, the location of the point of interaction between incoming neutrino and the target, in terms of which plane in the detector, is the desired output. 
For the hadron multiplicity problem, we count the number of out-coming charged hadron tracks with sufficient energy to traverse about two planes of the detector from the interaction.
A sample input image is given in Fig. \ref{fig:minerva_input}.
The networks were trained using data simulated by state-of-the-art physical models. 
In order that the networks are insensitive to differences between simulated and real images, some of the network populations were trained with a \emph{domain adversarial} component (DANN) \cite{JMLR:v17:15-239, Perdue_2018}.
For this work, we studied two separate output populations of vertex-finding networks and one population of hadron-multiplicity networks, each of which is based on 4,999 repetitions of the evolutionary algorithm.
In its running process, the algorithm was also allowed to alter layer types, layer order, and number of layers, as well as intra-layer features like kernel shapes, stride lengths, number of features, and type of non-linearity.
The data set of networks analyzed was thus built on a total of 299,050 unique network architectures.

All studies in this paper are reproducible using our analysis, extraction codes, and attributes data set, which are publicly available\footnotemark.
\footnotetext{https://github.com/Duchstf/CNN-Architectural-Analysis}The sample raw Caffe prototxt and output files from MENNDL are also documented with our codes\footnotemark[\value{footnote}].
Feature extraction codes were run using a SINGULARITY software container\footnotemark \cite{Singularity} for software deployment.\footnotetext{https://github.com/Duchstf/CNN-Architectural-Analysis-SingularityImg}

The rest of this paper is organized as follows. In Section \ref{Attribute_summary}, we describe several architectural attributes which were extracted from our set of networks.
In Section \ref{ML_results}, we present results of machine learning models built to predict CNN's performance based on the extracted architectural attributes.
Section \ref{attribute-analysis} details a possible way to get insights into CNN's behaviour relative to its architecture by analyzing the machine learning models' features.
Finally, in Section \ref{Conclusion}, we conclude by summarizing the study and discuss next steps required for further development of the research.

\section{Extracted Attributes Summary}\label{Attribute_summary}
Here we describe various network attributes which may be extracted and represented in a uniform way using a minimal amount of computation.
Several such attributes are the result of averaging over some groups of attributes.
This is because the size of groups of attributes may depend on the specific network architecture, and may not always serve the same functionality or be at the same scale in different networks and thus may produce ambiguity in interpretation.
For example, it is tempting to use network depth as an attribute, but different networks might have several input layers or several output layers.
In particular, some networks developed for analysis of MINERvA data all expect three input layers (corresponding to different angles of the same input image), and each produces two output layers (one for the domain classifier, and the other for the target classifier).
Thus, there is potential ambiguity in the notion of depth, since there are multiple input-to-output paths.
To remedy this issue, we can ask for the average depth.

Below is a list of all attributes extracted here.
Note that the list here is not exhaustive --- there are many other possibilities. 
Note that some attributes refer to ``horizontal'' and ``vertical'' directions.
Here we mean along the ``plane'' axis for horizontal and along the ``strip'' axis for vertical (Fig. \ref{fig:minerva_input}).
The abbreviations used in the analysis are given in [square brackets].

\begin{enumerate}
\item Average depth [\texttt{net\_depth\_avg}]
\item Number of convolutional layers [\texttt{num\_conv\_layers}]
\item Number of pooling layers [\texttt{num\_pooling\_layers}]
\item Average number of number of elements in outputs of fully-connected layers [\texttt{avg\_IP\_neurons}] 
\item Average number of connection parameters of fully-connected layers to previous layer [\texttt{avg\_IP\_weights}]
\item Average number of output feature maps in convolutional layers [\texttt{num\_conv\_features}]
\item Proportion of convolutional layers followed by a pooling layer [\texttt{prop\_conv\_into\_pool}]
\item Proportion of pooling layers followed by a pooling layer [\texttt{prop\_pool\_into\_pool}]
\item Proportion of convolutional layers with $1\times 1$ kernels [\texttt{prop\_1x1\_kernels}]
\item Proportion of convolutional layers with square kernel-shapes [\texttt{prop\_square\_kernels}]
\item Proportion of convolutional layers with horizontally-oriented kernels [\texttt{prop\_horiz\_kernels}]
\item Proportion of convolutional layers with vertically-oriented kernels [\texttt{prop\_vert\_kernels}]
\item Number of rectified linear unit (ReLU) activated convolutional layers [\texttt{num\_relu}]
\item Number of sigmoid-activated convolutional layers [\texttt{num\_sigmoid}]
\item Average percent reduction in activation grid area/ height/ width between consecutive convolutional layers \\\ [\texttt{avg\_grid\_reduction\_area/height/width
\_consecutive}]
\item Average percent reduction in activation grid area/ height/ width between input layers and final convolutional layers\\\ 
[\texttt{avg\_grid\_reduction\_area/height/\\\
width\_total}]
\item Proportion of convolutional layers using non-overlapping stride [\texttt{prop\_nonoverlapping}]
\item Average convolutional stride height/width \\\
[\texttt{avg\_stride\_h/w}]
\item Average ratio of convolutional layer's output feature maps to its depth [\texttt{avg\_ratio\_features\_to\_depth}]
\item Average ratio of layer's output feature maps to kernel area/height/width of convolutional layers \\\ [\texttt{avg\_ratio\_features\_to\_kerArea/Height\\\
/Width}]
\item Average ratio of kernel area/height/width to depth of convolutional layers \\\
[\texttt{avg\_ratio\_kerArea/Height/Width\_to\_depth}]
\end{enumerate}

Unravelling the attributes which can measure area, height, or width, this list comprises 32 architectural attributes.

\section{Predict CNNs' performance before training time.} \label{ML_results}
\subsection{Predictions for Vertex-Finding networks}
\subsubsection{Data summary}
We analyzed two populations of output networks designed for Vertex Finding in MINERvA using MENNDL.
For convenience, we refer to them as the \emph{First} and \emph{Second Populations}.
The first network in each genealogy in the two populations were initialized from the same set of networks.
However, they were optimized in separate MENNDL run-times, and trained based on different Caffe solver parameters. The first population was trained with a DANN component \cite{JMLR:v17:15-239, Perdue_2018}, whereas the second population was not.

\begin{figure}[btp]
    \centering
     \subfloat[]{%
        \includegraphics[width=0.5\linewidth]{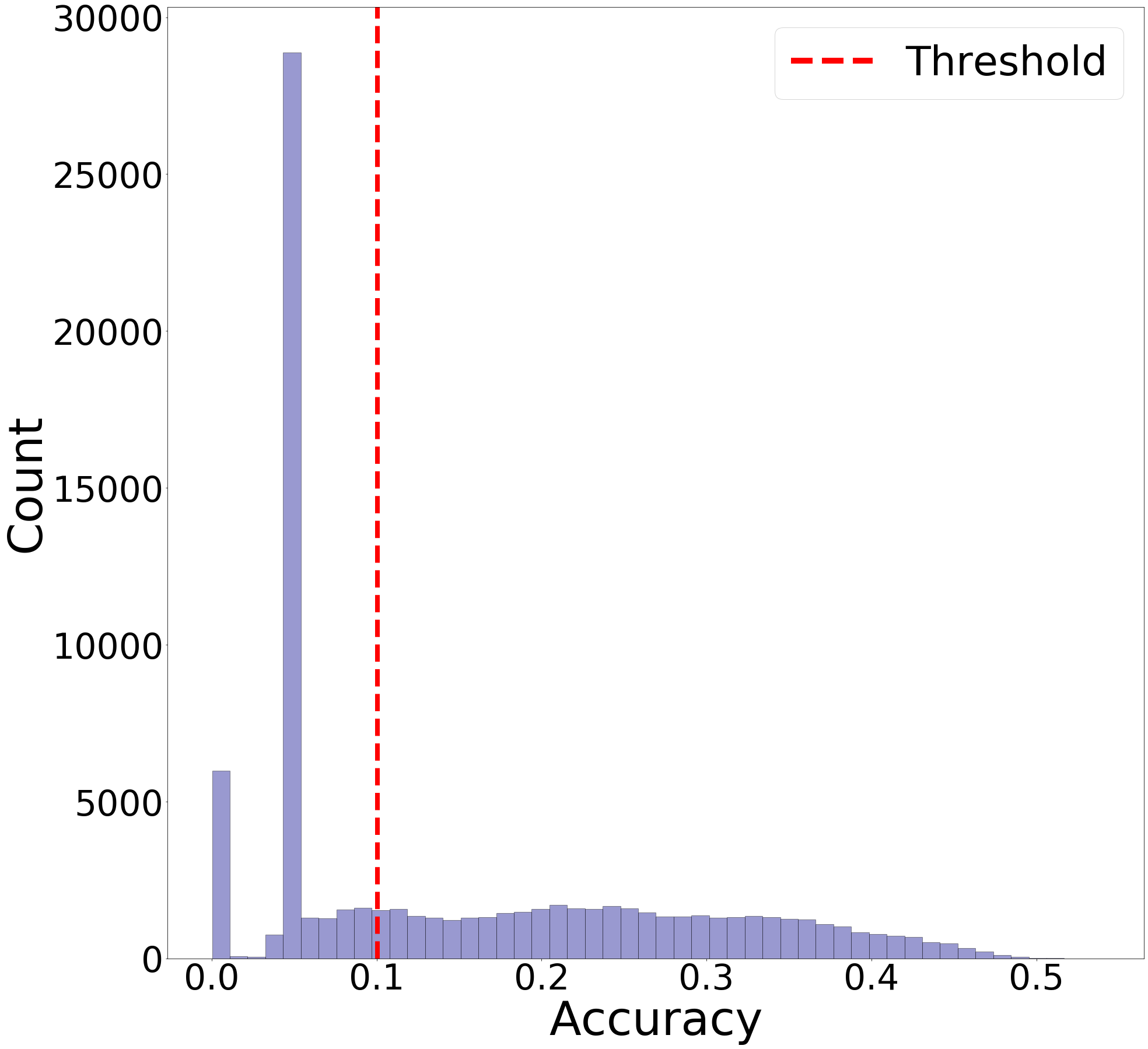}}
    \subfloat[]{%
        \includegraphics[width=0.5\linewidth]{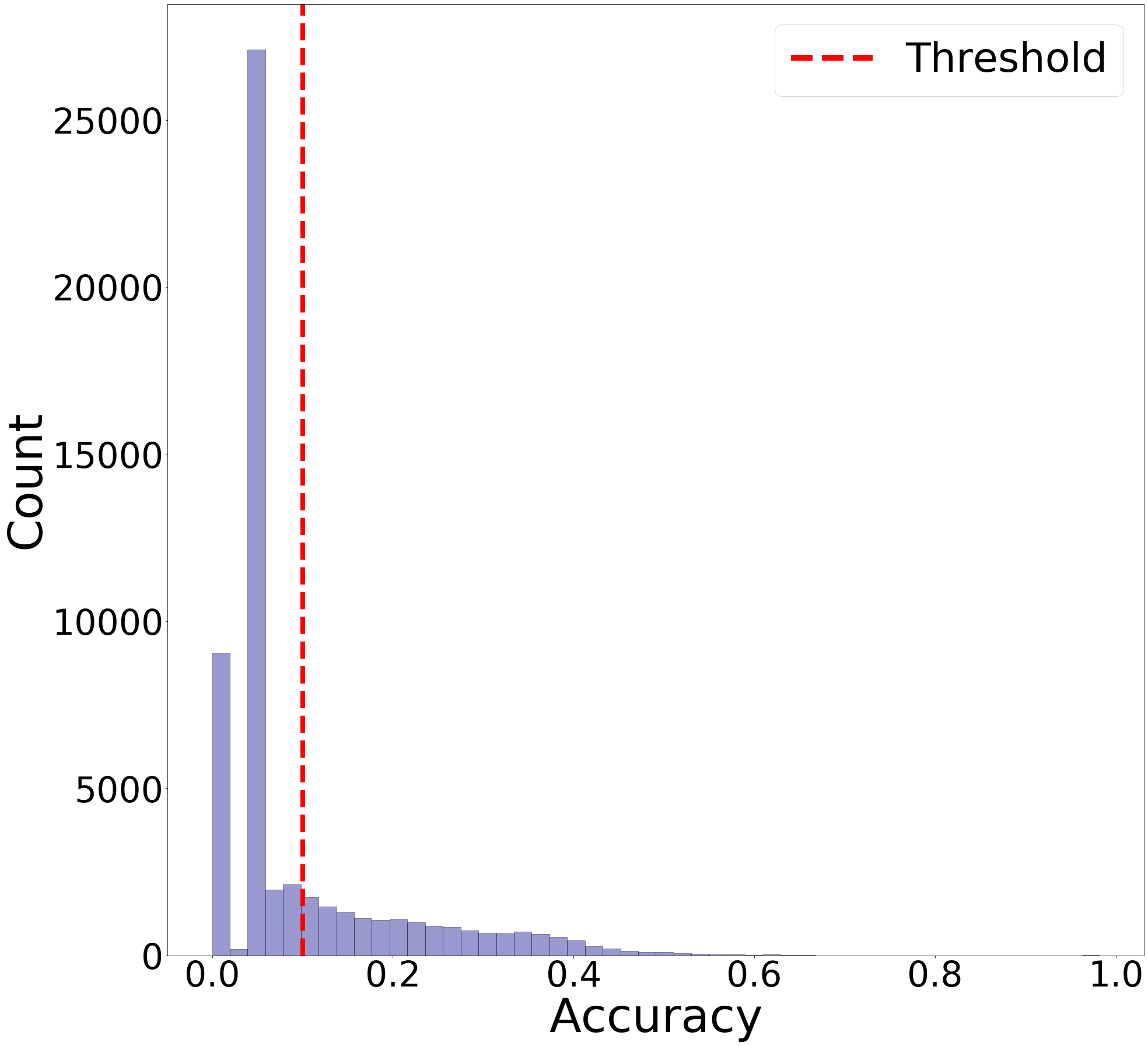}}
    \caption{Classification accuracy distributions for each population for vertex finding. Left: accuracy distribution of the first population. Right: accuracy distribution of the second population. The dotted lines represents where each dataset was divided into ``broken'' or ``healthy'' for our classification task.}
    \label{fig:Accuracy_dist_pop}
\end{figure}

In terms of accuracy, the networks have either 173 or 174 output classes corresponding to planes and targets in the detector. Therefore, the benchmark for random guessing is around $0.6 \%$. In Fig. \ref{fig:Accuracy_dist_pop}, both populations share very similar network accuracy distributions. 
They are both heavily left skewed with many networks' accuracy clustering around a very low value. 
For each population, we split the data into \emph{broken} and \emph{healthy} networks using threshold of $10.05 \%$, which is much higher than random guessing. 
The threshold was set so that the high peaks of very low performance network in the distributions are included in the \emph{broken} class, and the two classes are balanced.
The overall percentage of each category in each population is summarized in Table \ref{tab:percentage_broken_healthy}.

\begin{table}[bp]
\begin{center}
    \caption{Fraction of broken and healthy networks in each population.}
    \centering
    \begin{tabular}{|c|c|c|c|}
        \hline
        \textbf{Population} & \textbf{\textit{Broken}} & \textbf{\textit{Healthy}} & \textbf{Total number of networks} \\ 
        \hline
        First & 50.0\% & 50.0\% &83966\\ 
        \hline
        Second & 50.0\% & 50.0\% &31542 \\ 
        \hline
    \end{tabular}
    \label{tab:percentage_broken_healthy}
\end{center}
\end{table}

In this task, we choose to not combine the two populations together for fear that the mentioned inherent difference in the networks' attributes can interfere with our classification task and cause difficulties in interpreting the results.
For regression, we chose to combine the two populations together on the basis that we are only looking at the correlations of the network's attributes to predict the accuracy.

\subsubsection{Classification results}
Each population dataset was randomly split into training and testing sets with a 80/20 ratio, respectively.
Several algorithms were used to classify between between the two categories.
While more complex models, such as neural networks, were able to provide marginal improvements, we choose to only display results from Random Forest (RF) \cite{Breiman:2001:RF:570181.570182} and Extremely Randomized Tree (ERT) \cite{Geurts2006-ERT} for their performance and interpretability.
The algorithms and feature analysis are implemented using \texttt{scikit-learn} \cite{scikit-learn} library.

For this task, we propose a base accuracy of 50\%, since there is no class imbalance in both populations we used for classification.
The primary purpose of building machine learning models was to demonstrate the predictive nature of the architectural attributes, but not to perform further analysis based on the outputs of the models.

As can be seen in Table \ref{tab:vf_accuracy_score}, the scores are significantly better than random guessing (50\%), which underlines that the models were able to detect architectural separation between the attribute sets for \emph{broken} and \emph{healthy} networks.
Furthermore, the cross-validation scores and the accuracy on test set are very close together, so we would expect the models to have the same accuracy on unseen data set.

\begin{table}[tp]
\begin{center}
\caption{Accuracy of RF and ERT on train set and validation set in first \& second population.} \label{tab:vf_accuracy_score}
\begin{tabular}{|c|c|c|c|}
\hline
\textbf{Models} & \textbf{Population} & \multicolumn{2}{c|}{\textbf{Average accuracy scores}} \\ \cline{3-4} 
 &  & \textbf{Cross-validation} & \textbf{On test set} \\ \hline
RF & First & 67.3 $\pm$ 0.004\% & 66.8\% \\ \cline{2-4} 
 & Second & 69.6 $\pm$ 0.006\% & 70.7\% \\ \hline
ERT & First & 66.7 $\pm$ 0.006\% & 66.0\% \\ \cline{2-4} 
 & Second & 69.6 $\pm$ 0.007\% & 70.3\% \\ \hline
\end{tabular}
\end{center}
\end{table}

\subsubsection{Regression results}

After performing healthy/broken classification, we performed regression on the healthy networks in order to relate network features to the accuracy on the hold-out test set.
To prevent heteroscedasticity---where the sub-populations have different variabilities---in the data, the accuracies are transformed using the Box-Cox transformation \cite{Box-Cox64}.
The correlation between the independent variables and dependent variable remains the same after the transformation.
Before fitting, interaction terms between the original attribute set are also added.

\begin{table}[bp]
    \centering
     \caption{$R^2$ value of non-linear OLS model on individual populations and combined.}
    \label{tab:R_squared_pops}
    \begin{tabular}{|c|c|c|c|}
        \hline
        \textbf{Population} & \textbf{$R^2$} & \textbf{Adjusted $R^2$} & \textbf{Number of healthy networks}\\ 
        \hline
        First  & 0.445 & 0.439 & 41984 \\ 
        \hline
        Second  &  0.298 & 0.275 & 15771 \\ 
        \hline
         Combined  & 0.966 & 0.966 & 57755\\ 
        \hline
    \end{tabular}
\end{table}

\begin{figure}[tp]
    \centering
    \includegraphics[width = 0.48\textwidth]{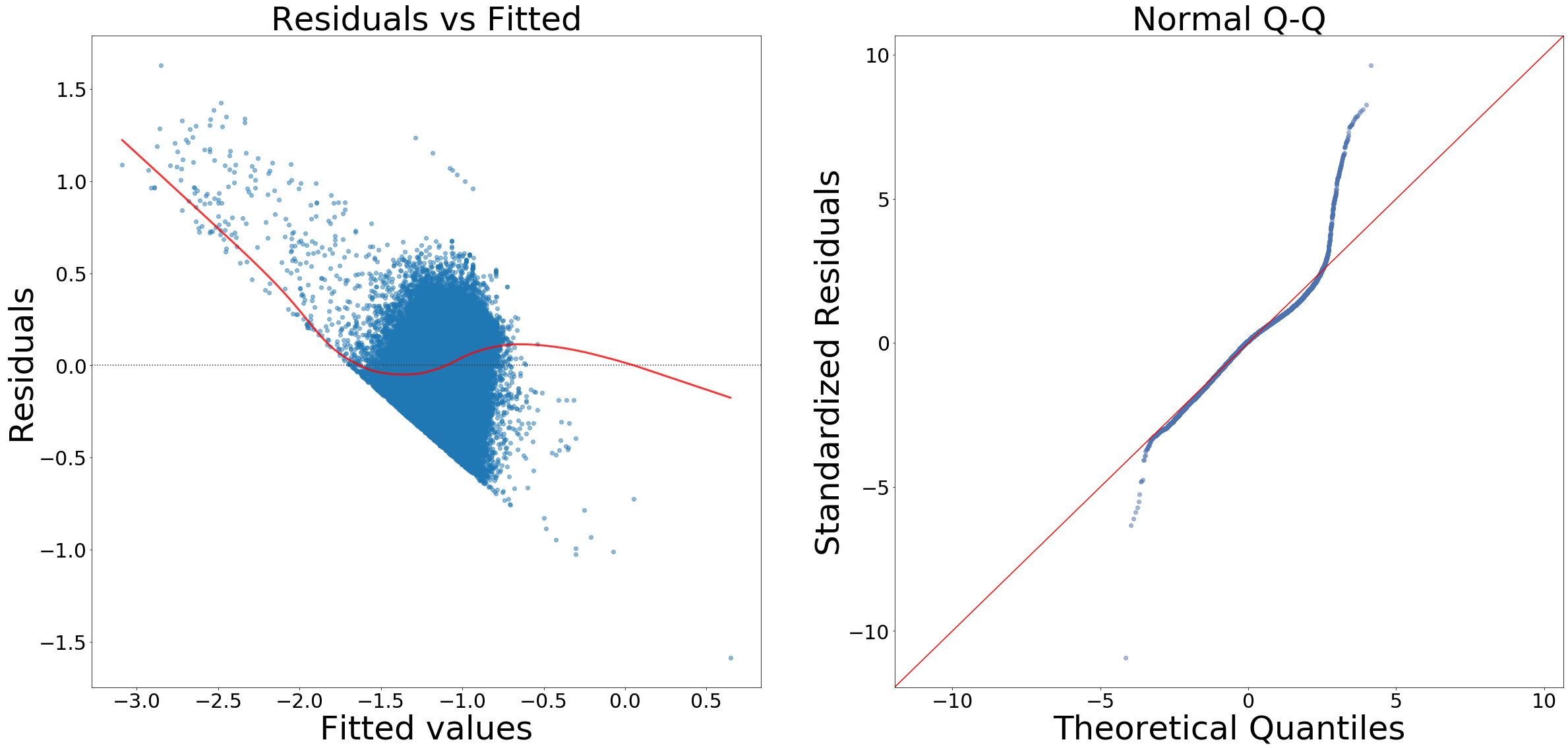}
    \caption{Residuals analysis of OLS model on the combined data set of healthy networks. Left: Scatter plot between residuals and the fitted value, indicating a linear relatioship between residuals and fitted. Right: Quantile-quantile plot depicting the distribution of standardized residuals -- the high tail indicates a gap in the residuals distribution.}
    \label{fig:Residual-OLS}
\end{figure}

Using a non-linear Ordinary Least Square (OLS) model with linear parameters, we performed regression separately on each population and then combine them together.
The OLS model is implemented using \texttt{StatsModels}\cite{statsmodels}.
As the two populations are distinct and we are looking merely at the relationship between network's architecture and its accuracy, combining them will not affect the regression process.
The results from the fit are summarized in Table \ref{tab:R_squared_pops}. In this case, while we cannot achieve a good $R^2$ for each individual population, the model is able to fit the combination of both populations.
A general trend is that as the number of networks increase, the $R^2$ value gets better.
This suggests that while we don't have enough events in the sub-populations to get a good fit, they overlap enough in the right regions of phase space to allow a good fit altogether.
However, it is worth noting that, as depicted in Fig. \ref{fig:Residual-OLS}, while the majority of residuals are distributed around 0, there seems to be a linear relationship between the residual and the fitted values, which means that more regressors are needed to account for this behaviour.
Furthermore, the Quantile-Quantile (Q-Q) plot in Fig. \ref{fig:Residual-OLS} with a high right tail indicates that there is a gap in the distribution of the residuals.
This is due to the fact that the accuracy's distribution is heavily left skewed with very few networks with high accuracy.

We also tried several regression algorithms that can account for a high level of non-linearity in the data. 
For example, we tried Decision Tree Regressors, Random Forest Regressors, Multi-level Perceptrons, Theil-Sen, and Huber regressors.
Almost all of them fail to generalize to validation data set and do not provide a significantly better $R^2$ than a simple OLS model.

\subsection{Prediction for Hadron-Multiplicity networks}
\subsubsection{Data summary}
\begin{figure}[bp]
    \centering
    \includegraphics[width = 0.3\textwidth]{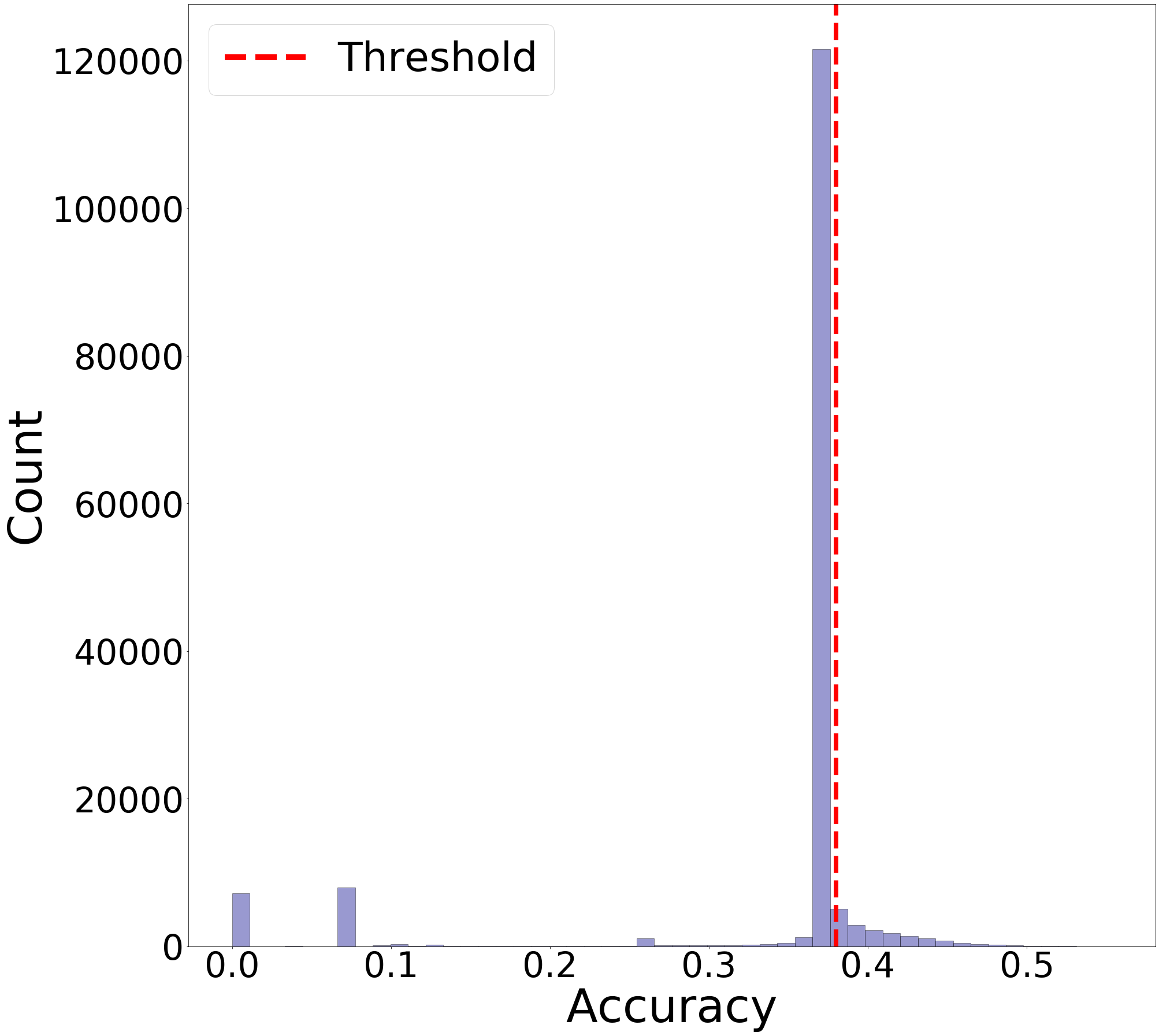}
    \caption{Accuracy distribution of the whole hadron multiplicity network population with threshold to divide between broken and healthy networks.}
    \label{fig:HM_Acc_Dist}
\end{figure}

Fig. \ref{fig:HM_Acc_Dist} depicts the accuracy distribution of hadron multiplicity networks with the threshold to divide to two classes of networks for classification.
To prevent class imbalances in the training data, we set the threshold to be 0.38 and broken networks were randomly sampled so that we have a 50/50 distribution between the two classes of 34614 networks in total. 

\subsubsection{Classification results}

For this task, we again used RF and ERT models to classify between broken and healthy networks.
The classification results are reported in Table \ref{tab:HM_Classification_accuracy}.
Both models consistently achieve accuracy of more than 70\% in both cross-validation on training set and testing set, which is 20\% better than random guessing (50\%), since there is no class imbalance.

\begin{table}[tp]
    \begin{center}
            \caption{Accuracy of RF and ERT on train set and validation set.}
    \label{tab:HM_Classification_accuracy}
    \begin{tabular}{|c|c|c|}
        \hline
        \textbf{Model} & \textbf{Average cross-validation score} & \textbf{Accuracy on test set}\\ 
        \hline
        RF  &  70.3 $\pm$ 0.006\% & 70.6\% \\ 
        \hline
        ERT  &  70.2 $\pm$ 0.003\% & 70.5\% \\ 
        \hline
    \end{tabular}
    \end{center}
\end{table}

Note that here we do not present regression's results for hadron multiplicity networks, since we have such a small amount of networks that the regression results are not significant to be presented.

\section{Attribute analysis}\label{attribute-analysis}
\begin{figure}[bp]
    \centering
        \subfloat{\includegraphics[width=0.5\linewidth]{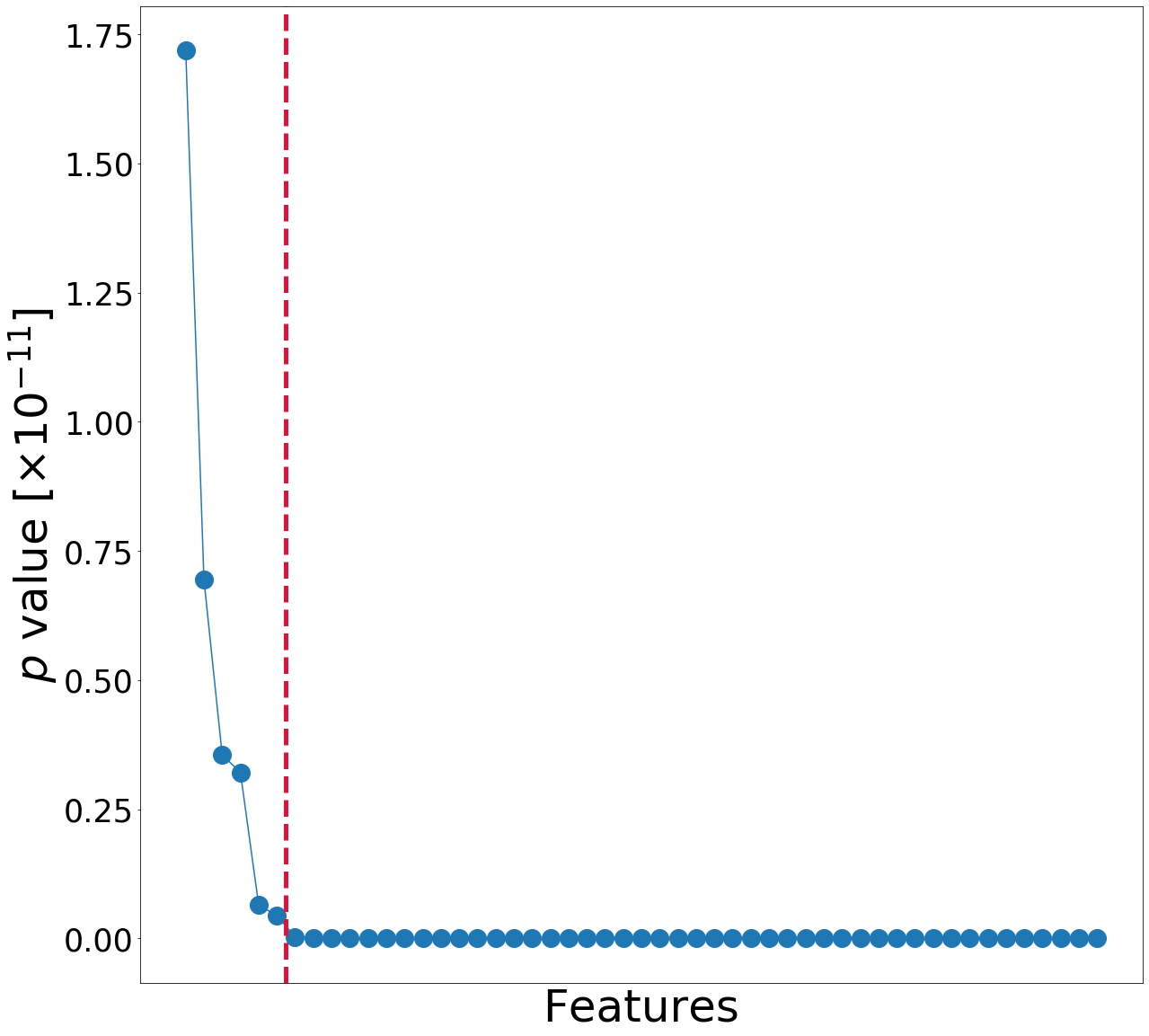}}
        \subfloat{\includegraphics[width=0.485\linewidth]{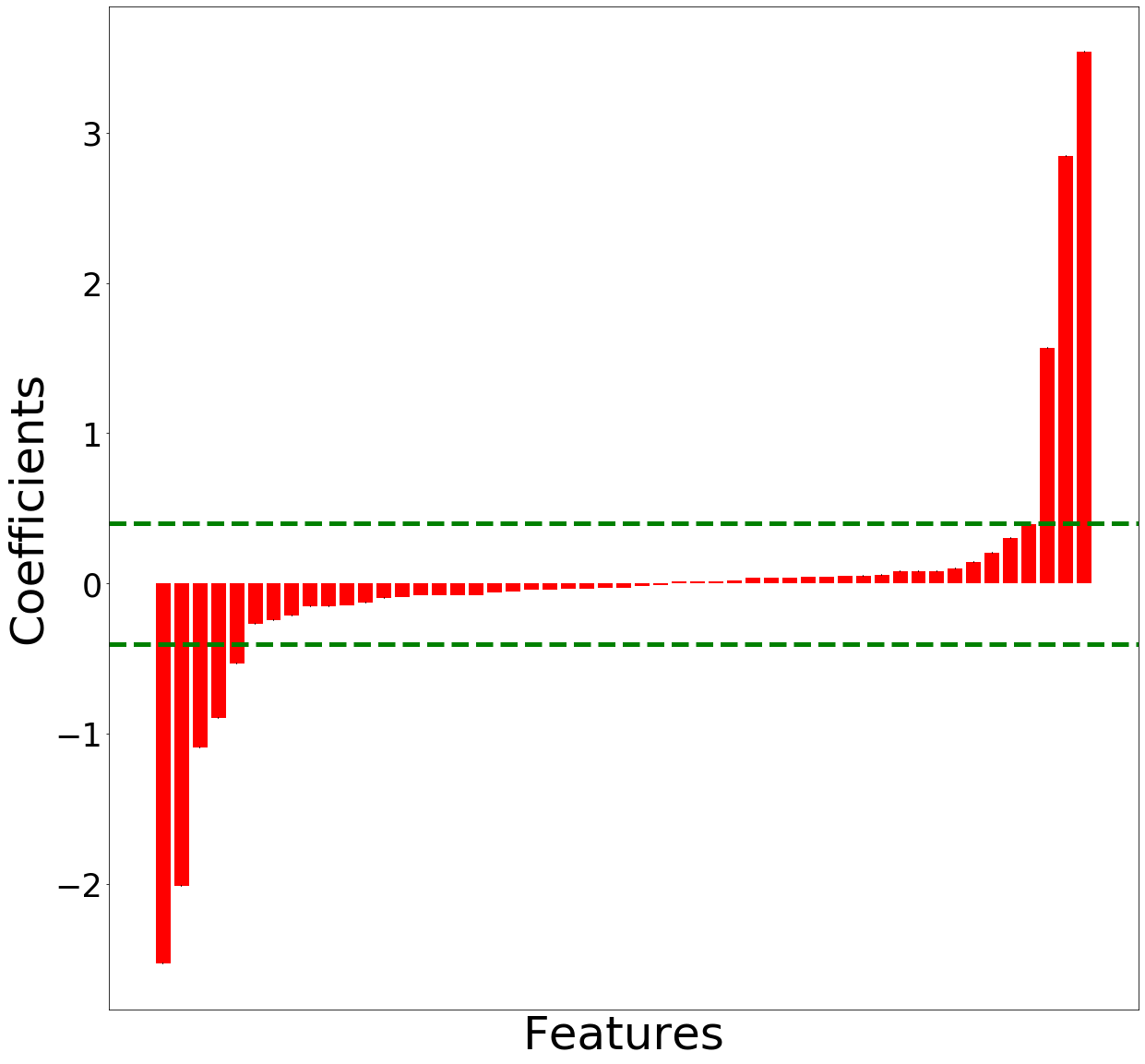}}
       \caption{Left: $p$ values of features that have $p < 2^{-10}$ plotted in descending order. Each dot represents the $p$ value of a feature. Several features have much more significant $p$ values than other features, which can be seen after the red dotted line. Right: coefficients of different features in descending order. Horizontal dotted green line represents the absolute 0.4 range within which many features' coefficients fall into. There are 8 features outside of this range that have significantly higher coefficients than the rest.}
    \label{fig:VF-p-value}
\end{figure}

Here we give some examples of how the attributes set can potentially be used to analyze the behaviour of the network's architecture.
While the OLS model for vertex finding networks cannot predict the accuracy of every network, the model has $p$ value under 0.05, and a significant $R^2$ value.
Many attributes have $p$ value under 0.05, and we observe that, when plotting the attributes' $p$ values that are less than $2^{-10}$ (Fig. \ref{fig:VF-p-value}:Left), there are 45 attributes and interactions that are much more important than the rest.
Since the data was normalized, the coefficients of variables are meaningful to look at.
We plot the coefficients of different features in Fig. \ref{fig:VF-p-value}:Right.
While many features have coefficients within 0.4 range from 0, there are 8 features that have significantly higher coefficients than the rest, which are reported in Table \ref{tab:OLS-params}. 

By looking at the values of the coefficients, we can make some insightful observations about the relationship of CNN's architecture and its accuracy in the vertex finding task.
\texttt{net\_depth\_avg}, \texttt{avg\_IP\_neurons} and their interactions are strongly correlated with the performance.
This suggests that increasing the capacity (number of parameters) of fully connected layers in the CNN can improve the overall performance of the CNN model.
Additionally, \texttt{num\_pooling\_layers} and \texttt{num\_conv\_layers} are negatively correlated with the performance.
This implies that, as we add more convolutional layers and pooling layers into the model, its performance will generally decrease. 
While the rest of the interactions are harder to interpret, the interaction term between \texttt{avg\_grid\_reduction\_height\_total} and \texttt{avg\_stride\_h} seems to point out an interesting property.
Typically in computer vision problems only square kernels are ever considered.
MINERvA physicists studied asymmetric kernel shapes for the vertex finding problem as a way of keeping the convolutions from reducing the image size along the planes axis \cite{Perdue_2018,ijcnn7966131}.

Overall, for the vertex finding and hadron multiplicity problem, our analysis of classification and regression models clearly indicates that it is possible to study a CNN model's accuracy prior to training by just looking at its architectural attributes. 
Moreover, analyzing the important features of the machine learning models can give us insights into how to potentially improve a CNN model's performance. 
That being said, our set of attributes is not extensive enough to fully characterize the complex relationship between CNN's architecture and its accuracy.
Further study of this is guaranteed.

\begin{table}[tp]
\begin{center}
\caption{Attributes that have significantly larger coefficients than those of other attributes in OLS model, as depicted in Fig. \ref{fig:VF-p-value}.}\label{tab:OLS-params}
\begin{tabular}{|c|c|}
\hline
\textbf{Variable} & \textbf{Coefficient} \\ \hline
net\_depth\_avg & $3.5 \pm 0.03$\\ \hline
avg\_IP\_neurons & $2.8 \pm 0.02$\\ \hline
avg\_IP\_neurons*net\_depth\_avg & $1.6 \pm 0.01$\\ \hline
avg\_grid\_reduction\_height\_total*avg\_stride\_h & -0.5 $\pm$ 0.02\\ \hline
avg\_IP\_neurons*num\_conv\_layers & -0.9 $\pm$ 0.01\\ \hline
avg\_IP\_neurons*num\_pooling\_layers & -1.1 $\pm$ 0.01\\ \hline
num\_conv\_layers & -2.0 $\pm$ 0.02\\ \hline
num\_pooling\_layers & -2.5 $\pm$ 0.02\\ \hline
\end{tabular}
\end{center}
\end{table}




\section{Summary and Outlook} \label{Conclusion}
In this paper, we proposed a systematic method that can be useful for uniform comparison of different architectural attributes of CNNs.
We demonstrated the predictive nature of those attributes in two specific problems---vertex finding and hadron multiplicity counting in MINERvA---through building machine learning models that predict the CNN's performance before its training time.
The classification models perform significantly better (66\% - 70\%) than random guessing (50\%). 
We were also able to achieve a significant OLS model with $R^2$ of  0.966 on a very large sample of Vertex-Finding networks.
Additionally, we detailed a potential method to study a CNN's behaviour relative to its architecture by analysing the predictive models' features. 

For future work, we plan to extend the architectural attributes set and take into account other hyper-parameters related to input domains and training process.
As we mentioned in the introduction, we did not look at anything related to the training process such as learning rate, momentum, optimization methods, etc.
We also did not study the learned weights and biases of the networks.
Additionally, statistics of the image dataset should be important (feature sizes and shapes, intensity distributions, etc.), although these can be challenging to quantify.
Considering some or all of the above might provide us with a more comprehensive study of network performance. 
We also want to have architectural attributes that account for more recent types of neural layers in the literature, e.g. \cite{2015arXiv151203385H,2016arXiv160806993H,2019arXiv190406952E}.
Furthermore, it can be interesting for us to perform the same kind of analysis on state-of-the-art network architectures and see to what extent does our current set of architectural attributes correctly characterize the network's performance. It is also promising to incorporate machine learning models such as the ones we built in this paper into model selection algorithms to evaluate a network's accuracy before training time, thereby boosting the efficiency of the algorithms.

\section*{Acknowledgment}

We would like to thank the MINERvA collaboration for access to their simulated data sets for this analysis.
MINERvA uses the resources of the Fermi National Accelerator Laboratory (Fermilab), a U.S. Department of Energy, Office of Science, HEP User Facility. 
Fermilab is managed by Fermi Research Alliance, LLC (FRA), acting under Contract No. DE-AC02-07CH11359, which included the MINERvA construction project. This material is based upon work supported by the U.S. Department of Energy, Office of Science, Office of Advanced Scientific Computing Research, Robinson Pino, program manager, under contract number DE-AC05-00OR22725. This research used resources of the Oak Ridge Leadership Computing Facility at the Oak Ridge National Laboratory, which is supported by the Office of Science of the U.S. Department of Energy under Contract No. DE-AC05-00OR22725. 


%


The US government retains and the publisher, by accepting the article for publication, acknowledges that the US government retains a nonexclusive, paid-up, irrevocable, worldwide license to publish or reproduce the published form of this manuscript, or allow others to do so, for US government purposes. 
The DOE will provide public access to these results of federally sponsored research in accordance with the DOE Public Access Plan (\url{http://energy.gov/downloads/doe-public-access-plan}).

\bibliographystyle{./bibliography/IEEEtran}
\bibliography{./bibliography/biblio}

\end{document}